# From Motion to Behavior: Hierarchical Modeling of Humanoid Generative Behavior Control

Jusheng Zhang[1], Jinzhou Tang[1], Sidi Liu[1],
Mingyan Li[1], Sheng Zhang[2], Jian Wang[3], Keze Wang[1,*]

[1]Sun Yat-sen University
[2]University of Maryland, College Park
[3]Snap Inc.

[*]Corresponding author: `kezewang@gmail.com`

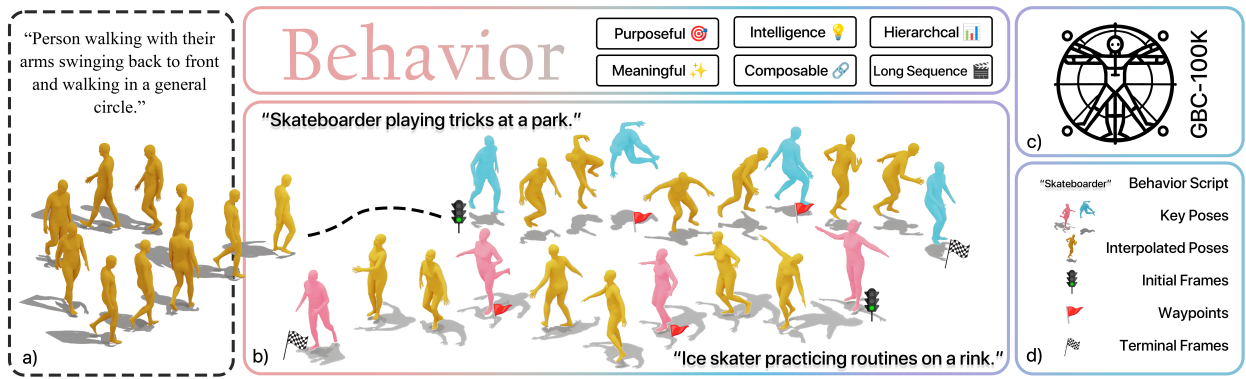

Figure 1. **From motion to behavior.** (a) Simple periodic motion patterns without complex, behavioral semantic meaning, (b) Complex, semantically meaningful human behaviors, demonstrating our framework's ability to generate goal-oriented, coherent behavior sequences. (c) Our proposed Generative Behavior Control (GBC) framework bridges this gap between low-level motions and high-level behavioral understanding.

## Abstract

*Human motion generative modeling or synthesis aims to characterize complicated human motions of daily activities in diverse real-world environments. However, current research predominantly focuses on either low-level, short-period motions or high-level action planning, without taking into account the hierarchical goal-oriented nature of human activities. In this work, we take a step forward from human motion generation to human behavior modeling, which is inspired by cognitive science. We present a unified framework, dubbed Generative Behavior Control (GBC), to model diverse human motions driven by various high-level intentions by aligning motions with hierarchical behavior plans generated by large language models (LLMs). Our insight is that human motions can be jointly controlled by task and motion planning in robotics but guided by LLMs to achieve improved motion diversity and physical fidelity. Meanwhile, to overcome the limitations of existing benchmarks, i.e., lack of behavioral plans, we propose GBC-100K dataset annotated with a hierarchical granularity of semantic and motion plans driven by target goals. Our experiments demonstrate that GBC can generate more diverse and purposeful high-quality human motions with 10× longer horizons compared with existing methods when trained on GBC-100K, laying a foundation for future research on behavioral modeling of human motions. Our dataset and source code will be made publicly available.*

## 1. Introduction

Recent advances in human motion analysis have been driven by deep learning innovations across multiple fronts.

3D pose estimation and tracking techniques [68, 80] now provide reliable foundations for behavior understanding. Moreover, motion generation has achieved realistic human movement synthesis [13, 20, 96], while physics-informed approaches [52, 89] ensure physical plausibility. Large language models (LLMs) further expand these capabilities [32, 77], demonstrate promising potential in high-level action planning and understanding.

Existing approaches face three key challenges: (1) maintaining temporal coherence over long durations, (2) ensuring physical plausibility across complex action sequences, and (3) bridging the semantic gap between high-level goals and low-level actions. Although short-term motion synthesis has shown considerable progress [13, 20, 51, 94, 96], existing methods struggle to maintain temporal coherence and physical plausibility over longer durations [26, 49, 66]. Physics-informed control policies [29, 52, 53, 74, 85, 87, 89] have improved motion stability. However, they often fail to bridge the gap between low-level actions and high-level goals [65, 72, 76]. In robotics, Task and Motion Planning (TAMP) approaches [10, 22, 44, 48, 59] attempt to integrate task planning with motor control, despite the remaining limitations in capturing human motion complexity [97]. These limitations reveal a critical gap: An exclusive focus on motion generation neglects the goal-oriented semantics, intentions, and planning inherent in human behavior. This insight motivates us to shift from motion generation to behavior generation, aiming to create extended sequences that are both physically feasible and semantically coherent to behavioral semantics.

To address these challenges, we introduce a novel task called Generative Behavior Control (GBC), which advances beyond traditional motion generation by focusing on two critical challenges: (1) generating complex, long-horizon action sequences that current methods struggle to model and (2) bridging action planning, motion generation, and intention understanding in a unified framework. Unlike conventional approaches [50, 94] that concentrate exclusively on motion synthesis, GBC aims to generate physically consistent **behaviors** that embody both semantic understanding and purposeful planning over extended sequences. While existing TAMP methods in robotics [10, 22, 44, 48, 59] typically emphasize precise geometric planning and deterministic action sequences, GBC addresses the inherent complexity of human-like behavior through probabilistic generation and semantic understanding. To enable this advancement, we develop GBC-100K, a large-scale dataset that addresses two fundamental challenges: (1) generating detailed, semantically aligned action sequences through fine-grained textual annotations, and (2) hierarchical mapping of high-level to low-level textual descriptions that facilitates behavioral planning with LLMs, as in Table 1.

Building upon these foundations, we present PHYLO-MAN (PHYsics-informed generative modeling of LOng-horizon planning for hierarchical huMANoid behaviour), a unified framework that makes two key technical contributions, as shown in Figure 1: (1) it integrates LLMs for high-level, goal-oriented action planning [1, 6, 9, 16, 34, 79] with TAMP-based robotics framework to ensure physical constraints [17, 29, 52, 54], and (2) it enables the generation of sequences that are 10× longer than traditional state-of-the-art motion generation methods. Specifically, PHYLOMAN realizes this behavior-centric approach through three main technical components: (1) a hierarchical planning framework that systematically decomposes high-level instructions into executable motion sequences through the integration of PoseScripts [15] and MotionScripts [88], (2) a parallel motion generation pipeline that simultaneously synthesizes multiple semantic-aligned movements with temporal consistency [12, 13, 15, 28, 39, 96], and (3) a physics-informed refinement mechanism based on control policy in a simulator [52, 89] that guarantees physical realism across extended sequences.

To enable systematic evaluation of our PHYLOMAN and facilitate research in behavior generation, we introduce GBC-100K, a large-scale multimodal dataset with ∼100K annotated video-SMPL pairs. The dataset addresses current limitations in temporal coherence and semantic complexity by providing hierarchical, fine-grained textual descriptions for generative models. Experimental results on both GBC-100K and HumanML3D [25] demonstrate PHYLOMAN's effectiveness in generating physically consistent and semantically coherent behaviors, showing significant improvements in temporal stability and action fidelity over extended sequences compared to existing methods.

Our **main** contributions are as follows:

- We introduce GBC-100K, a large-scale multimodal dataset of individual human behaviors with multi-level textual annotations, bridging the gap between high-level behavior understanding and low-level motion synthesis.
- We propose PHYLOMAN, a hierarchical framework that unifies LLM-driven planning with physics-based motion control, enabling coherent humanoid behaviors that are 10× longer than previous methods.
- Extensive quantitative and qualitative evaluations demonstrate that our PHYLOMAN achieves significant improvements in both physical consistency and semantic fidelity for extended behavior sequences compared to the state-of-the-art methods.

## 2. Related Work

**Behavior Decomposition.** In the field of behavior modeling, psychology, and sociology have laid foundational insights by decomposing complex human actions into fundamental components, exploring the influence of cognitive processes and societal structures [36, 58]. However, translating these theories into computational models remains challenging due to the complexity of mental states, social interactions, and environmental factors influencing human behavior [57]. Building on foundational insights from psychology and sociology, recent advancements in video understanding and action recognition have begun translating complex human behaviors into computational models by leveraging computer vision techniques, such as MotionLLM [83, 93], Video-LLaVA [46] and MiniGPT4-Video [3]. These models capture intricate temporal dependencies and relational context in sequential actions, which is particularly beneficial for instructional video analysis [70]. Here, methods that recognize both the hierarchical structure and the procedural flow of tasks have enabled a more nuanced understanding of human actions in real-world scenarios [4, 42]. Despite this progress, current methods still struggle with generating lifelike coherence and physical plausibility across continuous sequences [30, 50, 92, 94]. Addressing these limitations, our work aims to generate human behaviors that not only execute realistically in physical environments but also align with high-level semantic instructions, thereby advancing adaptability and realism in human behavior modeling.

Table 1. **Comparison of Existing Motion-Language Benchmarks.** We comprehensively compare our proposed GBC-100K with widely adopted human generation benchmarks across multiple dimensions (list in columns left to right): the total number of human action sequences (#Seq.), whether is based on SMPL-parameterized model [60] (SMPL) or/and video frames (Video), the total length of all videos (Len.), incorporation of hierarchical textual motion descriptions (Hierarchical), the number of distinct n-gram of words (Distinct-n@1) and phrases (Distinct-n@2) [45], average length of the texts(Avg. Len.), whether with goal-oriented (Goal orient.) textual annotations, whether can support open-vocabulary (Open Vocab.) motion synthesis. Our proposed GBC excels at long-horizon, fine-grained descriptions and diversity, with a large enough scale dataset of ∼100k SMPL clips and corresponding multi-level textual annotations.

| Datasets | #Seq. | SMPL | Video | Len. | Textual Annotations | | | | Goal Orient. | Open Vocab. |
|---|---|---|---|---|---|---|---|---|---|---|
| | | | | | Hierarchical | Distinct-n@1 ↑ | Distinct-n@2 ↑ | Avg. Len. ↑ | | |
| KIT-ML [61] | 3.9K | ✓ | ✗ | 10.3h | ✗ | 0.88 | 0.86 | 8.43 | ✓ | ✓ |
| UESTC [37] | 25.6K | ✗ | ✓ | 83h | ✗ | 0.71 | 0.90 | 2.58 | ✗ | ✗ |
| NTU-RGB+D [67] | 114.4K | ✗ | ✓ | 74h | ✗ | 0.69 | 0.88 | 3.12 | ✗ | ✗ |
| HumanAct12 [24] | 1.2K | ✓ | ✗ | 6h | ✗ | 0.73 | 0.89 | 1.97 | ✗ | ✗ |
| BABEL [62] | - | ✓ | ✗ | 43.5h | ✗ | **0.90** | 0.81 | 1.43 | ✗ | ✗ |
| HumanML3D [25] | 14.6K | ✓ | ✗ | 28.5h | ✗ | 0.46 | 0.86 | 12.37 | ✗ | ✓ |
| HMDB51 [41] | 6.8K | ✓ | ✓ | 7.8h | ✗ | 0.89 | 0.80 | 1.29 | ✗ | ✗ |
| COIN [70] | 46.3K | ✗ | ✓ | 476h | ✓ | 0.56 | 0.87 | 4.92 | ✗ | ✓ |
| ActivityNet [7] | 2K | ✗ | ✓ | **648h** | ✓ | 0.33 | 0.76 | 13.48 | ✓ | ✓ |
| **GBC-100k** | **123.7K** | ✓ | ✓ | 250h | ✓ | 0.51 | **0.91** | **50.92** | ✓ | ✓ |

**Human Motion Synthesis** encompasses several core areas: pose estimation, motion generation, motion prediction, and the application of physical constraints. In pose estimation, methods such as TRAM [81] and AiOS [68] have advanced the field by integrating techniques such as tracking and SLAM [56] to capture global trajectories and detailed human motions from in-the-wild videos. For human motion generation, generative models conditioned on inputs such as text or audio have made significant strides in creating realistic short-term movements [31, 50, 64, 94]. However, these models often struggle with achieving semantic coherence and physical plausibility across extended sequences. Similarly, while advances in motion prediction have refined the accuracy of forecasting future movements, these models frequently overlook physical feasibility, leading to sequences that may disrupt the coherence of generated behaviors [84, 95]. The primary limitation of existing methods is the absence of a cohesive framework that can jointly handle high-level planning and enforce physical constraints over long sequences. Motivated by this gap, we aim to unify semantic planning with physically consistent execution to enable robust, long-horizon behavior generation.

**Motion Control for Robotics.** Recent advancements in motion control and planning have explored various paradigms, including Task and Motion Planning (TAMP) [22] and learning-based approaches [90]. TAMP integrates high-level task planning with low-level motion execution, enabling robots to perform complex tasks by considering both discrete actions and continuous movements [11]. However, traditional TAMP methods often rely on predefined models and may struggle with adaptability in dynamic or unstructured environments [19, 97]. To address these limitations, learning-based techniques, such as reinforcement learning (RL) [69], have been integrated into TAMP frameworks, allowing robots to learn motion policies that adapt to environmental changes. Despite these advancements, RL approaches inherently struggle with generalization across diverse contexts and typically lack mechanisms for incorporating instructions, restricting their effectiveness in instruction-driven tasks and further necessitating extensive computational resources during training [5, 40, 55]. Existing approaches to motion control aim to achieve human-like behaviors by balancing high-level task planning with detailed motion execution [33]. However, existing methods often fall short, be-cause of their inability to dynamically integrate high-level semantic goals with low-level physical feasibility across long-horizon tasks [18]. Addressing this gap, our PHYLOMAN seeks to unify planning and control within a physics-informed framework, promoting coherent, adaptable behavior over extended sequences for the further application of embodied intelligence.

## 3. Preliminary

### 3.1. Generative Behavior Control

Generative Behavior Control (GBC) aims to synthesize long-term humanoid behaviors under both physical constraints and high-level semantic objectives. Specifically, GBC requires the generation of continuous motion sequences that span multiple minutes while maintaining both physical feasibility (e.g., joint limits, skate, float, and contact constraints) and semantic alignment with high-level behavioral instructions. Unlike traditional motion generation [50, 94], which primarily addresses short-term dynamics, GBC focuses on capturing the intentionality and semantic coherence of human behavior over long durations. This framework seeks to bridge the gap between low-level motor actions and overarching goals, facilitating the generation of more realistic, goal-oriented humanoid behaviors for applications in dynamic and complex environments.

In GBC, behavior is formally defined through a hierarchical script structure. A BehaviorScript $\mathcal{B} = \{\mathcal{D}, \mathcal{P}, \mathcal{A}\}$ consists of a high-level description $\mathcal{D}$, a set of PoseScripts $\mathcal{P} = \{p_0, p_1, ..., p_n\}$, and a set of MotionScripts $\mathcal{A} = \{a_0, a_1, ..., a_{n-1}\}$. The high-level description $\mathcal{D}$ serves as an abstract summary that encapsulates the overall semantic characteristics of the entire sequence of PoseScripts and MotionScripts. It follows a structured template with five key semantic elements: [Subject], [Emotion/State/Style], [Action], [Direction/Goal], and [Environment/Background] (e.g., "a person **energetically** dancing in circles at a party"). Each PoseScript $p_i$ describes a single atomic action (e.g., "raise right arm" or "stand in fifth position"), while each MotionScript $a_i$ captures a transition between consecutive poses $p_i$ and $p_{i+1}$, forming the complete sequence $(p_0, a_0, p_1, a_1, ..., p_{n-1}, a_{n-1}, p_n)$. For example, a BehaviorScript with $\mathcal{D}$ describing "an excited person dancing at a party" might decompose into MotionScripts like "spin energetically" and PoseScripts such as "hold an upbeat pose", where each component inherits and reflects aspects of the abstract description while providing concrete action details. Please refer to Supp. 7 for more details.

### 3.2. Task and Motion Planning

To effectively realize the proposed PHYLOMAN framework, we build upon Task and Motion Planning (TAMP), a hybrid planning paradigm that integrates the discrete decision-making of task planning with the continuous constraints of motion planning to generate feasible action sequences for robots operating in complex environments. Task planning involves reasoning over symbolic variables, such as selecting optimal movement sequences, coordinating multi-step locomotion, or generating physical gestures to convey intent. Motion planning, on the other hand, focuses on computing collision-free paths and physically valid trajectories in the robot's configuration space. The seamless interaction between these two levels is essential to bridge the gap between abstract planning and physical execution.

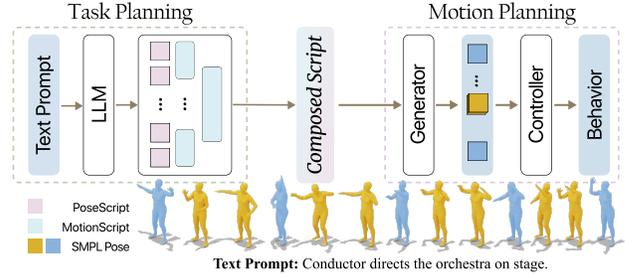

Figure 2. **Introduction of our PHYLOMAN Framework.** To address Generative Behavior Control, our PHYLOMAN decomposes behavior generation into two primary stages: Task and Motion Planning (TAMP). In Task Planning, a text prompt (e.g., 'Conductor directs the orchestra on stage') is processed by an LLM to generate a hierarchical Composed Script, consisting of keyframe-based PoseScripts interleaved with transition-focused MotionScripts. This structured representation ensures temporal coherence and behavioral fidelity. The Motion Planning stage utilizes a Generator composed of VAE and diffusion models to transform the Behavior Script into SMPL pose parameters, which are then refined by a physics-informed Controller to produce natural human behaviors. We demonstrate how PHYLOMAN enables fine-grained control over generated behaviors while maintaining motion naturalness and semantic alignment with the input prompt.

Formally, the TAMP problem can be represented as finding a sequence $(x_0, a_0, x_1, a_1, \ldots, x_T)$, where $x_i \in \mathcal{X}$ are robot configurations, and $a_i \in \mathcal{A}$ are actions. Each action must satisfy both task constraints $f(x, a) = \text{True}$ and motion feasibility constraints $g(x_i, x_{i+1}) \leq 0$. The task planner operates in a symbolic space, often represented using formal languages such as PDDL, determining which sequence of actions can achieve a given goal. The motion planner, on the other hand, operates in the continuous configuration space and computes paths that connect these symbolic actions while avoiding obstacles and respecting physical limitations.

## 4. Methodology

**Problem Formulation.** We formulate Generative Behavior Control (GBC) as a hierarchical planning problem that transforms high-level language instructions into physically executable SMPL poses. The problem consists of two hierarchical levels: task-level planning and motion-level planning.

At the task level, given an input prompt, the planner generates a sequence of PoseScripts $\{p_i\}_{i=0}^n$ and MotionScripts $\{a_i\}_{i=0}^{n-1}$. Each PoseScript $p_i$ maps to a configuration $x_i \in \mathcal{X} \subseteq \mathbb{R}^{J \times 3}$, where $\mathcal{X}$ denotes the set of physically valid joint orientations in the SMPL space, and $J$ represents the number of joints. Each motion script $a_i \in \mathcal{A}$ specifies a transition between configurations $x_i$ and $x_{i+1}$, where $\mathcal{A}$ defines the set of valid motion primitives. The task-level planning must satisfy:

$$C_T(x_i, a_i, x_{i+1}) = 0, \quad \forall i \in \{0, \ldots, n-1\} \quad (1)$$

where $C_T$ enforces a soft constraint on both physical validity and semantic consistency. Specifically, $C_T$ ensures that (1) transitions between configurations respect joint limits and maintain biomechanical feasibility, and (2) MotionScripts align with the intended behavior specified in the high-level instruction.

At the motion level, for each configuration pair $(x_i, x_{i+1})$, we compute a continuous trajectory $\tau_i : [0, 1] \to \mathcal{X}$ that satisfies the boundary conditions:

$$\tau_i(0) = x_i, \quad \tau_i(1) = x_{i+1}$$

and maintains physical constraints:

$$C_M(\tau_i(\lambda)) \leq 0, \quad \forall \lambda \in [0, 1] \quad (2)$$

where $C_M$ encompasses joint limits, collision avoidance, and dynamic feasibility constraints in simulation or real-world to ensure natural and physically plausible motion.

**Model Architecture.** To address this problem, we propose **PHYLOMAN**, a hierarchical planning framework that enables interpretable control over complex behaviors while bridging the gap between linguistic instructions and physical execution. The framework implements both task-level and motion-level planning through specialized components. Notably, PHYLOMAN operates as a pure inference framework where individual components are trained separately through stage-wise optimization.

The task planner integrates an LLM-based behavior planner with a conditional VAE. Given a high-level instruction (e.g., "conduct an orchestra on stage"), it generates a sequence:

$$(p_0, a_0, p_1, a_1, \ldots, p_{n-1}, a_{n-1}, p_n)$$

where each PoseScript $p_i$ is transformed into its SMPL representation $x_i$ through generation. The MotionScripts $a_i$ define transitions between poses, with the constraint $C_T(x_i, a_i, x_{i+1}) = 0$ enforced through a combination of LLM reasoning and its learned physical priors.

The motion planner combines a motion in-betweening model with an control policy to generate trajectories. For each transition specified by $(x_i, x_{i+1})$ and $a_i$, it produces a discrete approximation:

$$\{\tau_i(k\Delta\lambda)\}_{k=0}^K, \quad \Delta\lambda = \frac{1}{K}$$

of the continuous trajectory $\tau_i(\lambda)$, while maintaining $C_M(\tau_i(\lambda)) \leq 0$. This ensures smooth, physically valid transitions that respect joint limits, balance, and collision constraints.

Finally, a control policy (e.g., MPC or RL-based methods) tracks the entire trajectory from $x_0$ to $x_n$, interpolating between waypoints using the normalized parameter $\lambda$ to ensure that the generated motion sequence maintains physical feasibility by consistently enforcing $C_M(\tau_i(\lambda)) \leq 0$ throughout execution. A sliding window approach enables real-time performance while preserving both local stability and global task coherence. Please refer to Supp. 9 for more details.

**Data Collection.** Our data collection and organization process, outlined in Figure 3, consists of several stages. Initially, we collected raw videos from various internet sources. Specifically, we sampled ∼200k videos from the People & Society category in YouTube-8M and curated an additional ∼300k videos from datasets such as Kinetics, UCF101, HMDB, HM3.6M, and ActivityNet, resulting in around 500k raw videos. We then filtered these videos using YOLOv8-Pose to retain only those containing discernible human activities. To achieve this, we calculated a video score based on the number of visible humans per frame and the extent of body occlusion. Videos with scores below a defined threshold were discarded. Subsequently, we employed TransNetv2 to segment each video into clips, ensuring each clip contains no scene transitions. Long sequences were split into multiple clips, resulting in a final dataset of approximately 300k clips.

For data processing, each clip was subjected to both Motion Estimation and Motion Description. For Motion Estimation, we applied an advanced 3D Human Pose Estimation model, *i.e.* TRAM, to extract a sequence of SMPL representations across the clip. These SMPL sequences were then converted into PoseScript representations using predefined mapping rules. Moreover, we filtered out low-quality SMPL sequences based on several metrics, including physical plausibility, pose smoothness, and frame completeness. For Motion Description, we used a state-of-the-art multimodal large language model to generate a behavior script for each full video, structured as "[Subject] [Emotion/State/Style] [Action] [Direction/Goal] [Environment/Background]." Each clip from the video was then annotated to produce a corresponding MotionScript. Details on dataset-specific processing methods are provided in Supp. 10.3.

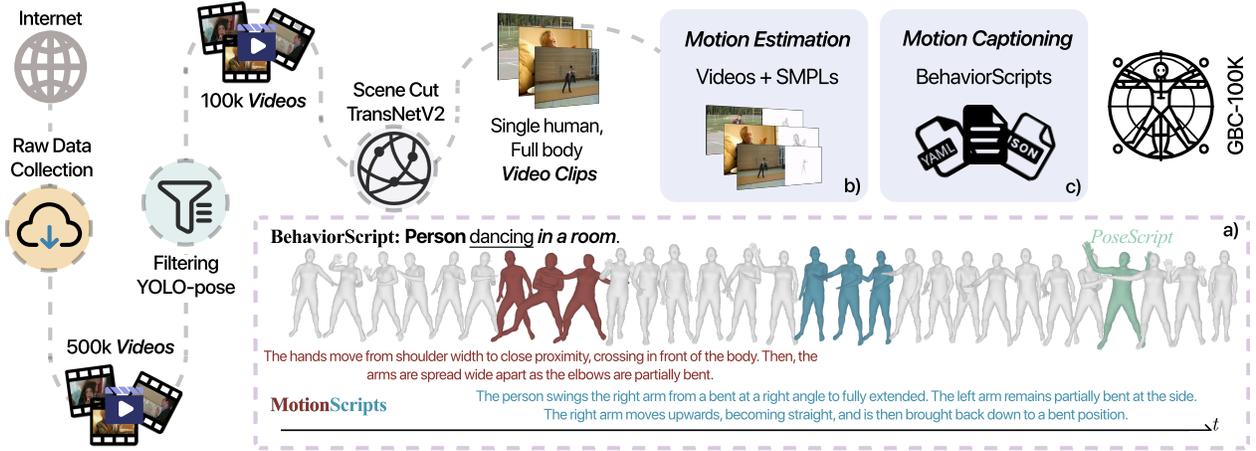

Figure 3. **Overview of collection process of our proposed GBC-100k dataset.** We begin by collecting large-scale raw videos from the internet and filtering them to retain only clips featuring a single, full-body person. Next, we apply advanced motion estimation and captioning techniques to extract human motions and generate textual annotations from the filtered clips, as illustrated in (a). Finally, we structure our data in the format shown in (b), organized hierarchically with a behavioral framework that includes (1) high-level behavior scripts, (2) detailed motion and PoseScripts, and (3) corresponding SMPL sequences.

### 4.1. Instructional Human Behavior Planning

**LLM as Behavior Planner.** We leverage LLMs as behavior planners for their unique strengths in common-sense reasoning and context understanding. This enables decomposing complex behavioral descriptions into structured, physically-plausible action sequences.

Building on established theories that human behavior is inherently hierarchical [43], we decompose behaviors temporally into keyframes (discrete postures) and transitions (inter-keyframe movements). We design structural primitives that encode both keyframe postures ($p_i$) and transitional motions ($a_i$), incorporating human body priors to ensure physical consistency.

Given a natural language behavior description $\mathcal{B}$, LLMs generate sequences of these primitives:

$$\{p_i\}_{i=0}^{n}, \{a_i\}_{i=0}^{n-1} = \text{LLM}(\mathcal{B}),$$

Through a naive Chain-of-Thoughts framework illustrated in Figure 2, we further enrich these sequences with kinematic attributes such as motion amplitude and speed.

**Parameterizing Text to Poses.** To reformulate GBC as a TAMP problem, an executable, parameterized configuration space is crucial for calculating plausible motion paths. However, a significant semantic gap between linguistic annotations and discrete postures complicates the direct mapping from text to pose. To address this challenge, we leverage text-conditioned generative models (*i.e.*, VAEs) to parameterize the projection from text to pose, ensuring semantic alignment between language and generated posture.

In particular, we adopt a pre-trained VAE from PoseScript [15], utilizing its text encoder $E_\phi$ and pose decoder $D_\theta$ to achieve diverse, flexible, and semantically aligned pose generation. For each input textual description $p_i$, the VAE encodes it into a latent representation $z_i$ through $E_\phi$, which is subsequently decoded by $D_\theta$ to generate the corresponding pose $x_i$:

$$z_i \sim E_\phi(p_i), \quad x_i \sim D_\theta(z_i), \quad \forall i \in \{1, \ldots, n\}.$$

### 4.2. Hierarchical Motion Generative Control

In this section, we introduce a hierarchical motion planner that leverages a generated sequence of pose-action pairs to guide motion in-betweening and continuous motion planning, producing the final coherent and physically consistent motion sequence.

**Motion Diffusion Prior.** In our PHYLOMAN, the motion diffusion prior $M \subseteq \mathbb{R}^{K,J,3}$ is generated using Diffusion Models [31, 94], where $K = N + C(N-1)$ and $C$ denotes the number of interpolated frames between consecutive configurations $x_i$. These models capture the complex distribution of human motion by iteratively refining samples from a Gaussian distribution, effectively modeling uncertainty in human pose transitions. Specifically, given a sequence of SMPL representations $\{x_i \in \mathcal{X}\}_{i=0}^{N}$ derived from a text-to-pose generator, this sequence serves as the input for the conditional diffusion model $\varphi_\theta$, which synthesizes an initial discrete trajectory connecting consecutive frames:

$$\mathcal{M} = \varphi_\theta(\mathbf{z}; \mathcal{X}, \mathcal{A}, t),$$

where $\mathbf{z} \in \mathbb{R}^{K,J,3}$ is a noise vector sampled from a Gaussian distribution and $t$ denotes the diffusion timesteps.

**Continuous Motion Planning.** Using $M$ as the reference, continuous motion planning produces the continuous trajectory $\mathcal{T}$ that ensures physical feasible transitions between each configuration $x_i$ and $x_{i+1}$ in a simulated environment. By leveraging a physics-based humanoid controller, *e.g.* PHC [52] and HOVER [30], the motion plan-ning process outputs control signals that guide the simulated avatar to follow the discrete poses from $M$, ensuring smooth and biomechanically accurate transitions. This approach refines the trajectory within the simulation, adapting to physical constraints and environmental dynamics, thereby producing a coherent, real-time motion sequence that aligns with the intended high-level behavior.

Table 2. Evaluating the SOTAs on **HumanML3D**, **GBC-10K**, and **GBC-100K** for Motion Generation Task. We use MotionScript and motion data only as the training set. Note that since all baselines cannot generate long sequences, we truncate GBC-100K into short sequences for evaluation. Besides, we use a balanced data mixture (*i.e.*, GBC-10K adapted with HumanML3D-style text descriptions + HumanML3D) in 2nd setup for fair comparisons. Additionally, we standardized the output length of all baseline models to 196 frames per motion sequence.

| Experiment | Method | R-Precision ↑ | | | FID ↓ | MM Dist ↓ | Diversity → | MultiModality ↑ |
|---|---|---|---|---|---|---|---|---|
| | | Top 1 | Top 2 | Top 3 | | | | |
| Trained & Evaluated on GBC-100K | Real | $0.501^{\pm.007}$ | $0.743^{\pm.002}$ | $0.833^{\pm.002}$ | $0.003^{\pm.001}$ | $2.426^{\pm.008}$ | $5.980^{\pm.104}$ | - |
| | MotionLCM [13] | $0.497^{\pm.003}$ | $\mathbf{0.751}^{\pm.002}$ | $\mathbf{0.827}^{\pm.005}$ | $0.816^{\pm.032}$ | $2.743^{\pm.003}$ | $3.846^{\pm.079}$ | $\mathbf{3.391}^{\pm.063}$ |
| | MDM [73] | $0.417^{\pm.004}$ | $0.539^{\pm.003}$ | $0.629^{\pm.008}$ | $0.387^{\pm.118}$ | $2.657^{\pm.101}$ | $2.452^{\pm.162}$ | $2.207^{\pm.091}$ |
| | MotionCLR [8] | $0.527^{\pm.003}$ | $\mathbf{0.751}^{\pm.005}$ | $0.838^{\pm.004}$ | $0.114^{\pm.000}$ | $2.472^{\pm.009}$ | $4.364^{\pm.000}$ | - |
| Trained & Evaluated on HumanML3D+GBC-10K | Real | $0.497^{\pm.004}$ | $0.663^{\pm.002}$ | $0.706^{\pm.003}$ | $0.005^{\pm.002}$ | $3.726^{\pm.006}$ | $7.266^{\pm.023}$ | - |
| | MotionLCM | $0.485^{\pm.006}$ | $0.648^{\pm.005}$ | $0.663^{\pm.007}$ | $0.641^{\pm.009}$ | $3.314^{\pm.006}$ | $7.723^{\pm.016}$ | $3.212^{\pm.082}$ |
| | MDM | $0.307^{\pm.004}$ | $0.478^{\pm.006}$ | $0.655^{\pm.005}$ | $0.296^{\pm.008}$ | $4.725^{\pm.003}$ | $7.407^{\pm.017}$ | $2.139^{\pm.082}$ |
| | MotionCLR | $0.537^{\pm.002}$ | $0.692^{\pm.006}$ | $0.761^{\pm.008}$ | $0.161^{\pm.000}$ | $3.314^{\pm.006}$ | $2.364^{\pm.000}$ | - |
| Trained & Evaluated on HumanML3D | Real | $0.511^{\pm.003}$ | $0.703^{\pm.002}$ | $0.797^{\pm.002}$ | $0.002^{\pm.002}$ | $2.794^{\pm.008}$ | $9.503^{\pm.065}$ | - |
| | MotionLCM | $0.502^{\pm.003}$ | $0.703^{\pm.003}$ | $0.805^{\pm.002}$ | $0.467^{\pm.012}$ | $2.986^{\pm.009}$ | $9.631^{\pm.065}$ | $2.172^{\pm.082}$ |
| | MDM | $0.320^{\pm.002}$ | $0.505^{\pm.004}$ | $0.607^{\pm.005}$ | $0.544^{\pm.044}$ | $\mathbf{2.452}^{\pm.162}$ | $9.559^{\pm.068}$ | $2.799^{\pm.072}$ |
| | MotionCLR | $\mathbf{0.542}^{\pm.001}$ | $0.733^{\pm.002}$ | $\mathbf{0.827}^{\pm.002}$ | $\mathbf{0.099}^{\pm.003}$ | $2.981^{\pm.006}$ | $2.145^{\pm.043}$ | - |

## 5. Experiments

**Experimental Setup** The primary objective of our experiments is to validate the hypotheses concerning the efficacy of our proposed method in generating extended human motion sequences that maintain behavioral continuity while adhering to high-level directives. Our PHYLOMAN aims to enhance the dataset's quality and granularity, yielding improved realism and detailed motion outputs. Furthermore, we conduct comparative experiments to evaluate the contributions of goal orientation, intentionality, and social dynamics within our behavior planning strategy. We conduct ablation studies to assess the impact of individual model components on overall performance.

**Evaluation Metrics.** Our comprehensive evaluation employs a suite of metrics: Multimodal Distance (MM Dist), Diversity, Success Rate (SR), Physical Error (PhysErr), R-Precision, Fréchet Inception Distance (FID), Motion Length, and MultiModality. SR is assessed through human evaluation to gauge the practical effectiveness of generated behaviors. Detailed metric definitions and calculation methods are provided in Supp. 10.2. Additionally, the details of the user study for evaluating the SR value are listed in Supp. 10.4.

**Implementation Details.** We train PHYLOMAN with a batch size of 1024 over 100 epochs using Adam optimizer with an initial learning rate of $10^{-5}$. We apply a cosine annealing schedule to decay the learning rate to $10^{-3}$. The CLIP-based similarity metric is trained on our dataset to ensure domain-specific evaluation. Notably, the CLIP model and the diffusion model are trained on different splits of our dataset to ensure unbiased evaluation. Specifically, we sampled approximately 25k motion clips to fine-tune ActionCLIP, and 2k clips to fine-tune CondMDI [12] pre-trained on HumanML3D with T5 text encoder [63] for fine-grained, long-horizon linguistic conditioning. Our PHYLOMAN is implemented in PyTorch and all experiments are conducted on a single NVIDIA RTX-4090 GPU. The training time is about 6 hours per 20,000 samples, while the inference time is about 1 minute per sample with 1000 frames.

### 5.1. Comparative Benchmarking

In Table 2, we validate the quality of our proposed dataset by evaluating multiple state-of-the-art baseline methods (i.e., MotionCLR [8], MDM [73], MotionLCM [13], T2M-GPT [91], MoMask [27] CondMDI [12]) across three distinct configurations (GBC-100K, HumanML3D+GBC-10K, and HumanML3D). Please refer to Supp. 10 for more details. Since existing baselines cannot generate long sequences, we truncate GBC into shorter sequences for evaluation and employ a balanced data mixture for fair comparisons.

The experimental results demonstrate that all methods trained on GBC-100K achieve higher MultiModality scores, attributable to the fine-grained textual annotations in our dataset. This finding (1) validates the finer-grained and more semantically aligned motion descriptions in our dataset; and (2) evidences that GBC is more challenging and comprehensive than HumanML3D, closer to real-world be-haviors, which substantially benefits future research. While gaps exist in FID metrics, these can be attributed to our dataset's construction from internet videos, featuring more diverse motion ranges, varied video quality, and potentially less precise text alignment.

Table 3. Zero-shot Evaluation on GBC. To evaluate the effectiveness of PHYLOMAN and its individual components, we conducted comprehensive experiments using a textual dataset of 1,000 BehaviorScripts generated by GPT-4o, which served as the input for PHYLOMAN. We assessed our PHYLOMAN using metrics specifically designed for the GBC task to evaluate the model's performance in behavior generation and control. "Discard" denotes the corresponding component is disabled. "Heuristic" denotes the rule-based template matching strategy following [38].

| Component | Methods | Phys-Err ↓ | Diversity ↑ | Succ. Rate ↑ |
|---|---|---|---|---|
| Motion Generator | MoMask [27] | $0.224^{\pm.028}$ | $96.28^{\pm.089}$ | $0.328^{\pm.000}$ |
|  | T2M-GPT [91] | $0.131^{\pm.094}$ | $99.92^{\pm.281}$ | $0.179^{\pm.000}$ |
|  | Discard | - | - | - |
| LLM Planner | Heuristic | $0.141^{\pm.012}$ | $19.32^{\pm.017}$ | $0.118^{\pm.000}$ |
|  | Discard | $1.031^{\pm.094}$ | - | $0.067^{\pm.000}$ |
| Text-to-Pose | Heuristic | $0.101^{\pm.042}$ | $97.73^{\pm.411}$ | $0.452^{\pm.000}$ |
|  | ChatPose [21] | $0.293^{\pm.050}$ | $103.48^{\pm.239}$ | $0.613^{\pm.000}$ |
|  | Discard | - | - | - |
| Controller | PHC [52] | $0.105^{\pm.050}$ | $99.59^{\pm.447}$ | $0.793^{\pm.000}$ |
|  | Discard | $0.235^{\pm.076}$ | $101.5^{\pm.253}$ | $0.762^{\pm.000}$ |
| - | **Optimal** | $\mathbf{0.093^{\pm.039}}$ | $\mathbf{109.7^{\pm.253}}$ | $\mathbf{0.821^{\pm.000}}$ |

Table 4. Evaluation on the effectiveness of Hierarchical Annotations across Different LLM Planners. After fine-tuning on our GBC-100K dataset, our PHYLOMAN demonstrates significant improvements in behavior planning performance, even outperforming closed-source LLM models by a slight margin.

| Setting | Model | Diversity ↑ | MultiModality ↑ | Succ. Rate ↑ |
|---|---|---|---|---|
| Fine-tuned | Llama3.1-70B [23] | $105.37^{\pm.215}$ | $\mathbf{3.052^{\pm.020}}$ | $0.753^{\pm.000}$ |
|  | Qwen-V2.5-72B [86] | $\mathbf{112.24^{\pm.112}}$ | $2.721^{\pm.025}$ | $\mathbf{0.821^{\pm.000}}$ |
|  | DeepSeek-V3 [14] | $107.82^{\pm.419}$ | $2.629^{\pm.020}$ | $0.806^{\pm.000}$ |
| Zero-shot | Llama3.1-70B | $92.53^{\pm.226}$ | $2.224^{\pm.028}$ | $0.702^{\pm.000}$ |
|  | Qwen-V2.5-72B | $96.83^{\pm.128}$ | $2.103^{\pm.030}$ | $0.708^{\pm.000}$ |
|  | DeepSeek-V3 | $97.26^{\pm.722}$ | $2.157^{\pm.030}$ | $0.653^{\pm.000}$ |
|  | GPT-4o [35] | $103.84^{\pm.076}$ | $2.309^{\pm.030}$ | $0.807^{\pm.000}$ |
|  | Claude-3.5-sonnet [2] | $109.21^{\pm.093}$ | $2.251^{\pm.030}$ | $0.778^{\pm.000}$ |

## 5.2. Long-Sequence Behavior Generation

Table 3 presents our comprehensive ablation studies on long-sequence (1024 frames) motion generation, the central problem that most existing methods cannot address. Our PHYLOMAN is compared against two state-of-the-art adapted baselines (MoMask and T2M-GPT) and various ablated variants. we adopt the following components for optimal performance: CondMDI [12] serves as the Motion Generator; Chain-of-Thought [82] functions as the LLM Planner; the text-to-pose conversion method proposed by Ginger Delmas et al. [15]; and HOVER operates as the Controller.

The results unequivocally demonstrate PHYLOMAN's significant performance improvements across all key metrics: Compared to methods without TAMP, our PHYLOMAN achieves a 133% increase in the success rate (from 0.3 to 0.7) while reducing the physical error by 91% (from 1.224 to 0.105). These findings confirm that our proposed TAMP pipeline simultaneously achieves superior task completion, physical plausibility, and naturalness: the three pillars of high-quality motion generation. For detailed visual comparison, please refer to Supp. 10.5.

## 5.3. Analysis of Planning and Physical Constraints

**Hierarchical Planning.** We evaluate the effectiveness of our hierarchical planning framework through comprehensive quantitative and qualitative analyzes. As shown in Table 3, our LLM planner significantly outperforms Heuristic method, achieving substantially higher success rates (0.821 vs. 0.118) and greater diversity (109.7 vs. 19.32).

**Text-to-Pose Mapping.** Compared to heuristic approaches, our PHYLOMAN exhibits a marginal decrease in Physical Error (0.093 vs. 0.101) and achieves a substantial 81.6% improvement in Success Rate (0.821 vs. 0.452) and an 12.2% enhancement in Diversity (109.7 vs. 97.73), indicating a clear advantage in the completion of practical tasks.

**Simulator.** Our comparison between variants with and without physics simulation reveals that incorporating physics reduces Physical Error by 60.5% (from 0.235 to 0.093) while maintaining identical Success Rates (0.821) and marginally increasing Diversity (by 8.0%), underscoring the importance of physical constraints in preserving motion quality.

**Fine-tuning LLMs.** Our experimental analysis shows that fine-tuning an LLM on hierarchical annotations from our dataset significantly improves motion sequence quality. As shown in Table 4, fine-tuned LLMs outperform pretrained ones. The dataset, derived from internet videos, was annotated with motion and behavior scripts using a VLM. These annotations, encoding a structured semantic hierarchy, were used to fine-tune the LLM as a motion planner, enabling more coherent and contextually appropriate trajectories. The key insight is that internet videos inherently capture realistic human behavior patterns, providing rich semantic information. By integrating these patterns, we achieved notable performance gains, highlighting the value of real-world data and structured hierarchies in motion planning. This approach bridges raw video data with semantically rich motion generation, advancing computer vision applications.

## 6. Conclusion

In this paper, we introduce a physics-informed framework supported by a large-scale multimodal benchmark, designed for language-driven behavioral planning and physics-based motion control. This framework enables the generation of coherent, ultralong humanoid behaviors. Looking ahead, we aim to expand the framework for practical applications, including embodied intelligence and digital avatars.

# From Motion to Behavior: Hierarchical Modeling of Humanoid Generative Behavior Control

## Supplementary Material

## 7. Terminology

Here, we explain three key terms in our PHYLOMAN for readers unfamiliar with related topics.

**PoseScript.** The term "PoseScript" is used to describe the specific configuration of the human body at a given moment in time. This configuration is expressed through a description of the spatial characteristics of various body parts, including joint angles, inter-limb distances, relative positions, body orientations, and ground contact states.

**MotionScript.** The term "MotionScript" is used to describe the temporal characteristics of human movement over a period of time. It expresses the process of human movement over a period of time by describing the temporal characteristics of the movement of a body part, including, but not limited to, the direction of movement, amplitude, duration, and sequence of movement.

**BehaviorScript**. A BehaviorScript is utilized to comprehensively describe a human action by integrating multiple PoseScripts and MotionScripts into a cohesive high-level expression, such as "changing a tire on a bicycle." In particular, a complete BehaviorScript is comprised of a high-level abstract behavioral statement and a sequence of interleaved PoseScripts and MotionScripts that represent a set of static actions and the transitions between them. This definition enables the framework to generate transitions between each set of actions in parallel, as well as to modify each transition separately.

## 8. Notations

We summarize notations in this paper as follows:

| | |
|---|---|
| $\mathcal{X}$ | Configuration space of the SMPL model |
| $\mathcal{A}$ | Action space of MotionScripts |
| $\mathcal{T}$ | Trajectory space |
| $\mathcal{J}$ | Set of all joints |
| $\mathcal{K}$ | Set of all collision pairs |
| $x_{i,j}$ | Angle of joint $j$ at configuration $x_i$ |
| $\bar{v}_j, \bar{a}_j$ | Maximum allowable velocity and acceleration of joint $j$ |
| $\sigma(c, v)$ | Smooth penalty function: $\max(0, v-c)^2$ |
| $g(a_i)$ | Feature mapping function: maps action $a_i \in \mathcal{A}$ to the feature space |
| $\bar{g}$ | Expected semantic feature vector |
| $\kappa, \epsilon_s$ | Transition steepness ($\kappa$) and semantic tolerance threshold ($\epsilon_s$) |
| $d_k(\lambda)$ | Distance between collision pair $k$ at trajectory progress $\lambda$ |
| $\mathbf{M}(x)$ | Mass matrix for SMPL |
| $\mathbf{C}(x, \dot{x})$ | Coriolis forces for SMPL |
| $\mathbf{G}(x)$ | Gravitational force vector for SMPL |
| $\tau(\lambda)$ | Joint torques at trajectory progress $\lambda$ |

## 9. Behavior Constraints

In this section, we expound on the technical and theoretical details of our proposed approach in Section 4.1. To synthesize human motion that aligns with semantic expectations and physical feasibility, constraints are defined over three spaces: the configuration space of the SMPL [60] model $\mathcal{X}$, the action space of MotionScripts [88] $\mathcal{A}$, and the trajectory space $\mathcal{T}$. Two key constraints are proposed: the high-level transition constraint $C_T : \mathcal{X} \times \mathcal{A} \times \mathcal{X} \to \mathbb{R}$ and the low-level motion constraint $C_M : \mathcal{T} \to \mathbb{R}$.

### 9.1. High level constraints

The high-level constraint is defined as:

$$C_T(x_i, a_i, x_{i+1}) = w_1 f_j(x_i, x_{i+1}) + w_2 f_s(a_i),$$

where $w_1, w_2 > 0$ are weights. Here, $f_j(x_i, x_{i+1})$ enforces joint-based constraints, and $f_s(a_i)$ ensures semantic consistency.

The joint constraint $f_j(x_i, x_{i+1})$ combines range and biomechanical limits:

$$f_j(x_i, x_{i+1}) = w_a f_r(x_{i+1}) + w_b f_b(x_i, x_{i+1}),$$

where $w_a, w_b > 0$. The range constraint $f_r(x_{i+1})$ ensures each joint $j \in \mathcal{J}$ lies within anatomically plausible limits:

$$f_r(x_{i+1}) = \sum_{j \in \mathcal{J}} [\sigma(x_j^{\max}, x_{i+1,j}) + \sigma(x_{i+1,j}, x_j^{\min})],$$

where $x_j^{\min}$ and $x_j^{\max}$ represent the minimum and maximum allowable joint angles. The biomechanical constraint $f_b(x_i, x_{i+1})$ penalizes excessive joint velocities $\dot{x}_{i+1,j}$ and accelerations $\ddot{x}_{i+1,j}$:

$$f_b(x_i, x_{i+1}) = \sum_{j \in \mathcal{J}} \left[\sigma(\bar{v}_j, \|\dot{x}_{i+1,j}\|) + \sigma(\bar{a}_j, \|\ddot{x}_{i+1,j}\|)\right],$$

and $\sigma(c, v) = \max(0, v-c)^2$ penalizes values exceeding the limits $\bar{v}_j$ (velocity) and $\bar{a}_j$ (acceleration).

The semantic function $f_s(a_i)$ aligns action $a_i$ with expected semantic features:

$$f_s(a_i) = \frac{\|g(a_i) - \bar{g}\|^2}{1 + \exp(-\kappa(\|g(a_i) - \bar{g}\| - \epsilon_s))},$$

where $g(a_i)$ is obtained by combining a diffusion model, which generates SMPL sequences, with a CLIP-based SMPL encoder [78] that extracts semantic features from these sequences. The expected semantic feature $\bar{g}$ is obtained from ground-truth SMPL sequences using the same encoder. $\kappa > 0$ controls the transition steepness, and $\epsilon_s > 0$ is the semantic tolerance.

### 9.2. Low level constraints

For trajectory $\tau_i$, the low-level motion constraint aggregates joint limits, collision avoidance, and dynamic feasibility:

$$C_M(\tau_i) = w_3 g_j(\tau_i) + w_4 g_c(\tau_i) + w_5 g_d(\tau_i),$$

where $w_3, w_4, w_5 > 0$ balance the terms. Practically, these constraints are derived from the MuJoCo [75] simulator, ensuring realistic dynamics. In this study, MuJoCo models dynamics with discrete time steps ($\Delta t$) using semi-implicit Euler integration:

$$q_{t+\Delta t} = q_t + \Delta t \cdot v_{t+\Delta t}. \tag{3}$$

Contact forces are computed using implicit optimization methods, ensuring numerical stability during trajectory simulation. In this context, the joint path constraint $g_j(\tau_i)$ ensures limits along the trajectory:

$$g_j(\tau_i) = -\int_0^1 \sum_{j \in \mathcal{J}} \left[\sigma(x_j^{\max}, x_j(\lambda)) + \sigma(x_j^{\min}, x_j(\lambda))\right] d\lambda.$$

Collision avoidance $g_c(\tau_i)$ prevents violations of the minimum allowable distance $d^{\min}$:

$$g_c(\tau_i) = -\int_0^1 \sum_{k \in \mathcal{K}} \frac{\sigma(d^{\min}, d_k(\lambda))}{1 + \exp(-\kappa_c(d^{\min} - d_k(\lambda)))} d\lambda,$$

where $d_k(\lambda)$ represents the distance of the $k$-th collision pair. $\kappa_c > 0$ is the collision steepness parameter, which controls the rate at which the penalty increases as the distance $d_k(\lambda)$ approaches $d^{\min}$.

The dynamic feasibility constraints ensure that the synthesized motion trajectory adheres to the physical dynamics of the SMPL model. In Mujoco, the dynamics are governed by the equation of motion:

$$\mathbf{M}(x)\ddot{x} + \mathbf{C}(x, \dot{x}) + \mathbf{G}(x) = \tau,$$

The dynamic feasibility constraint is then formulated as:

$$g_d(\tau_i) = -\int_0^1 \|\mathbf{M}(x(\lambda))\ddot{x}(\lambda) + \mathbf{C}(x(\lambda), \dot{x}(\lambda)) + \mathbf{G}(x(\lambda)) - \tau(\lambda)\|^2 d\lambda,$$

where $\mathbf{M}$ is the mass matrix, $\mathbf{C}$ represents Coriolis forces, $\mathbf{G}$ is gravitational force, $\tau(\lambda)$ denotes joint torques generated through control policy, and $T > 0$ is the total duration of the motion trajectory.

## 10. Additional Experimental Details

### 10.1. Baselines

We evaluate our PHYLOMAN on a variety of baselines that achieve state-of-the-art performance in generative quality, diversity, and semantic alignment. We briefly introduce each baseline as follows:

- **T2M-GPT** [91]: Combines Vector Quantized Variational Autoencoders (VQ-VAE) with Generative Pre-trained Transformers (GPT) to produce high-quality motion sequences aligned with textual inputs.
- **MoMask** [27]: Employs a generative mask modeling framework with hierarchical quantization, using masked and residual transformers to generate multi-layered high-fidelity motions.
- **MDM** [94]: Utilizes a diffusion-based generative approach, generating motions through gradual denoising guided by textual descriptions.
- **MotionLCM** [13]: Learns latent representations of motion, enabling effective modeling of text-to-motion mappings in latent space.
- **MotionCLR** [8]: Applies contrastive learning to capture the correspondence between text and motion, ensuring the generated sequences align with textual inputs.
- **CondMDI** [12]: Introduces Flexible Motion In-betweening, capable of generating precise and diverse motions with flexible spatial constraints and text conditioning.

### 10.2. Evaluation Metrics

To comprehensively evaluate the proposed method, we adopt a range of metrics from MDM [94], MotionLCM [13] and PhysDiff [89]. Each metric has been carefully selected to capture different aspects of the generated motion sequences, such as their alignment with high-level textual directives, physical plausibility, and behavioral diversity. Our evaluations leverage established methodologies from prior works, ensuring consistency and comparability with existing benchmarks.

1. **Multimodal Distance (MM Dist):** Measures the alignment between generated motions and their corresponding textual descriptions. Leveraging ActionCLIP fine-tuned on GBC-100K, we extract feature embeddings for both the generated motions and their textual counterparts. The average cosine distance between these embeddings is computed to quantify alignment, with lower values indicating better correspondence.

2. **Diversity and MultiModality:** Diversity captures the variance of generated motions across the entire dataset by calculating the pairwise feature distances between randomly sampled motion sequences, following the definitions outlined in MotionLCM [13]. MultiModality, in contrast, measures the diversity of generated motions conditioned on the same textual description. This is achieved by sampling two subsets of motions for each textual description and averaging the pairwise distances between their feature embeddings. Together, these metrics reflect the richness and multi-modal nature of the generated outputs.
3. **Success Rate (SR):** Evaluates the practical utility of the generated motion sequences in completing intended tasks. To compute SR, we conducted a human evaluation study using, where participants were presented with generated motion sequences and their corresponding high-level directives. More details can be found in Supp. 10.4.
4. **Physical Error (Phys-Err):** Computed following the methodology from PhysDiff [89], it includes three components: ground penetration (Penetrate), floating violations (Float), and foot sliding (Skate). Penetrate measures the distance between the ground and the lowest mesh vertex below it, while Float measures the distance of the lowest mesh vertex above the ground. Skate quantifies the horizontal displacement of foot joints during ground contact in adjacent frames. A tolerance of 5 mm is applied to account for geometric approximations. Phys-Err is the aggregate sum of these components, providing a holistic measure of physical plausibility.
5. **Fréchet Inception Distance (FID):** Used to evaluate the quality of generated motions by comparing their feature distributions to those of ground-truth motions. FID is calculated by extracting embeddings using ActionCLIP and computing the Fréchet distance between the distributions of real and generated motions. Lower FID values indicate closer alignment between the two distributions.
6. **R-Precision:** Assesses text-motion alignment by measuring the proportion of correct matches between generated motions and ground-truth motions, given a textual description. For each description, the top-k closest motions in the embedding space are retrieved, and R-Precision is computed as the percentage of ground-truth motions among the retrieved sequences. This metric is consistent with the definitions used in MDM.

### 10.3. Prompt Template

In our PHYLOMAN, we implement a Chain-of-Thought (CoT) approach to decompose high-level behavior instructions into structured motion sequences. This process consists of two main stages:

**Behavior Understanding and Planning** First, we employ a large language model to comprehend the high-level instruction and generate a structured behavior plan. The model outputs:
- A concise summary capturing the core behavior category
- A detailed description explaining the timing, body movements, objectives, and interactions involved

**Behavior Understanding Prompt**

> You are an assistant designed to translate high-level instructions into a sequential behavior plan...
>
> Given the following instruction: "instruction"
>
> DO NOT output additional words or any code block but only the summary and description. Please generate a summary and a behaviour description, both in natural language, keep summary short with a few words.
>
> **Example Outputs:**
>
> ```
> {
>     "summary": "Short summary of the
>     instruction.",
>     "description": "Behaviour description
>     for the instruction."
> }
> ```

**Sequential Decomposition** The behavior plan is then decomposed into two complementary scripts using the following detailed prompt:

**Sequential Decomposition Prompt**

> You are an assistant that transforms high-level behavior instructions into a structured low-level, long-horizon motion sequence for single humanoids. Each element in the sequence contains both a 'keyframe' and a 'transition'. The 'transition' describes the action connecting this keyframe to the next one.
>
> **Output Format:**
>
> ```
> [
>     {
>         "keyframe": "Description of the
>     first pose or state.",
>         "transition": "Description of the
>     transition to the next keyframe."
>     },
>     ...
>     {
>         "keyframe": "Description of the
>     last pose or state.",
>         "transition": ""
>     }
> ]
> ```

**Rules for Keyframe:**
1. Identify Key Body Parts: Focus on arms, legs, head, torso
2. Use Defined Posecodes:
   - *Angle Posecodes:*
     - straight
     - slightly bent
     - partially bent
     - bent at a right angle
     - almost completely bent
     - completely bent
   - *Distance Posecodes:*
     - close
     - shoulder width apart
     - spread
     - wide apart
   - *Relative Position Posecodes:*
     - X-axis: 'at the right of', 'x-ignored', 'at the left of'
     - Y-axis: 'below', 'y-ignored', 'above'
     - Z-axis: 'behind', 'z-ignored', 'in front of'
   - *Pitch & Roll Posecodes:*
     - vertical
     - horizontal
     - pitch-roll-ignored
   - *Ground-Contact Posecodes:*
     - on the ground
     - ground-ignored
   - *Orientation Posecodes:*
     - X-axis: lying flat forward to lying flat backward
     - Y-axis: leaning left to leaning right
     - Z-axis: about-face turned clockwise to counterclockwise
   - *Position Posecodes:*
     - X/Y/Z-axis: significant left/downward/backward to right/upward/forward
3. Subject Selection: Identify the most active joint as the subject
4. Ensure descriptions indicate static posture, not dynamic motion

**Rules for Transition:**
1. Provide overview of human action
2. Use specified posecodes for describing changes
3. Include movement directions: Forward, Backward, Left, Right
4. Describe speed and magnitude of movements
5. Maintain temporal relationships between concurrent movements

**Example Outputs:**

*Keyframe:*
The person is standing upright with a slight forward lean. The left arm is slightly bent and extended outward. The right arm is bent at a right angle, with the hand positioned near the chest. The legs are straight and shoulder-width apart.

*Transition:*
The person moves far to the right. At the same time, he is moving way over forward at an average pace. A moment later, he turns clockwise. The left elbow is bent at right angle, from that pose, the left elbow is extending greatly and very fast.

This decomposition results in:

**a) PoseScript (Keyframes):** Each keyframe describes a static posture using standardized pose codes as detailed in the prompt rules.

**b) MotionScript (Transitions):** Each transition describes the dynamic motion between keyframes following the specified guidelines.

### 10.4. User Study

We conducted a comprehensive user study to evaluate the performance of our PHYLOMAN framework against existing motion generation baseline models. The study involved 20 participants with diverse backgrounds from our institution. Each participant was presented with 10 samples randomly selected from a sample pool containing 100 examples per method.

**Evaluation Metrics** We developed a Success Rate (Succ. Rate) metric that comprehensively evaluates the generated behaviors. This metric is normalized to [0,1] and is calculated as follows:

$$SR = w_1 C + w_2 Q \tag{4}$$

where $SR$ represents the final Success Rate, $C$ represents the completion score of necessary action steps, $Q$ represents the quality score, and $w_1 = 0.5$ and $w_2 = 0.5$ are the respective weights. The quality score $Q$ is further composed of multiple sub-metrics evaluated on a 5-point Likert scale:

- Motion fluidity ($F$)
- Body part coordination ($CO$)
- Natural rhythm ($R$)
- Transition smoothness ($S$)
- Action completeness ($AC$)
- Step completion ($SC$)
- Detail preservation ($D$)
- Text-motion alignment ($A$)

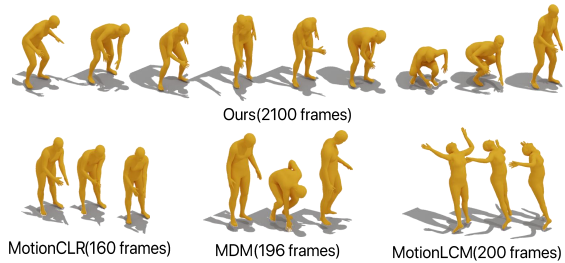

Figure 4. **Qualitative Comparison of Motion Generation Methods.** Visual demonstration of motion sequences generated for the textual prompt: *"A mechanic changes a tire on a bike in a garage."* Our proposed approach produces temporally extended sequences that better capture the complete action while maintaining semantic consistency with the textual description, outperforming baseline methods in both sequence length and motion quality.

These sub-metrics are combined using the following formula:

$$Q = \frac{F + CO + R + S + AC + SC + D + A}{40} \quad (5)$$

where each component is scored from 1 to 5, with:
- Score 1: Very poor/unsatisfactory
- Score 3: Average/neutral
- Score 5: Excellent/very satisfactory

**Survey Structure.** The questionnaire was designed to evaluate each sample on all eight aspects using a standardized 5-point Likert scale. For each metric, participants were provided with clear definitions.

### 10.5. Additional Visualizations

To further demonstrate the capabilities and limitations of PHYLOMAN, we present qualitative results including three successful cases and one failure case of long-horizon behavior generation in Figure 4, Figure 5 and Figure 6. The successful cases showcase: (1) Martial artist performs arts techniques, (2) Athlete training at a gym, and (3) Dancer rehearsing in a studio. These examples highlight our PHYLOMAN's ability to maintain semantic alignment and physical plausibility over long sequences, while successfully decomposing high-level behavioral goals into coherent motion sequences. In contrast, our failure case shows a scenario where a person attempts to perform swimming, where the model struggles to maintain physical balance during rapid transitions and exhibits temporal inconsistency in action sequencing. These visualizations collectively demonstrate both the strengths of our PHYLOMAN in handling structured, goal-oriented behaviors and its current limitations in extremely dynamic scenarios requiring precise physical coordination.

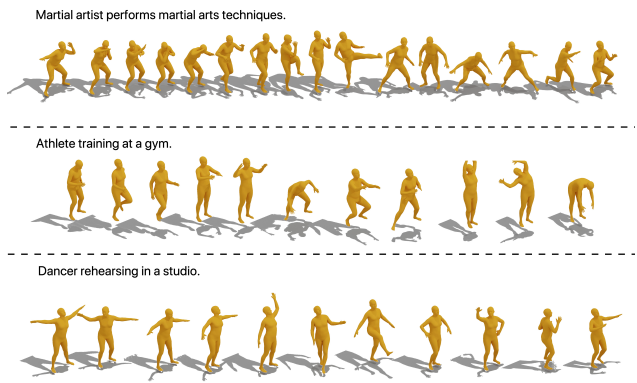

Figure 5. Successful samples.

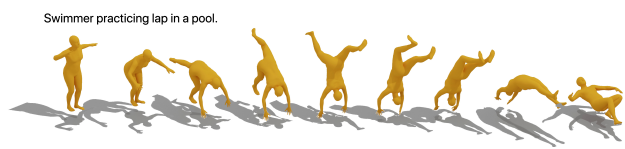

Figure 6. Failed sample.

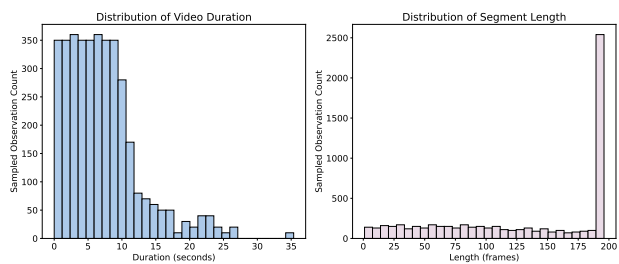

Figure 7. The duration statistics of sampled 3780 video clips and corresponding annotated motion segments with MotionScripts.

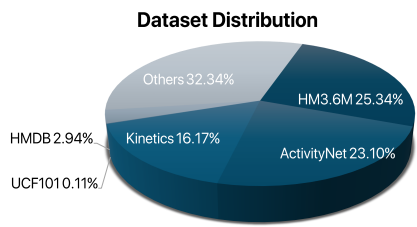

Figure 8. **Data composition in GBC-100k.** The dataset comprises contributions from multiple sources: HM3.6M (25.34%), ActivityNet (23.10%), Kinetics (16.17%), HMDB (2.94%), UCF101 (0.11%), and others (32.34%). The "Others" category includes curated subsets from the Motion-X [47] and FLAG3D [71] motion capture datasets, as well as selected videos from YouTube-8M.

## 10.6. Dataset Details

As shown in Figure 9, Figure 8, and Figure 7, our GBC-100k dataset is a large-scale, multimodal resource designed to support research on generative behavior control. It consists of diverse video-SMPL-text triplets, where each sample includes a video clip, its corresponding SMPL pose sequence, and textual descriptions in the form of BehaviorScripts, MotionScripts, and PoseScripts. The dataset integrates data from multiple well-known sources, including HM3.6M, ActivityNet, Kinetics, HMDB, UCF101, and curated sources such as Motion-X and FLAG3D, alongside selected videos from YouTube-8M, which collectively account for 32.34%. This broad composition ensures a wide range of human activities, providing diversity in both motion and context. Each source contributes to the richness of the dataset, making it a benchmark for tasks requiring fine-grained motion understanding and behavioral annotation.

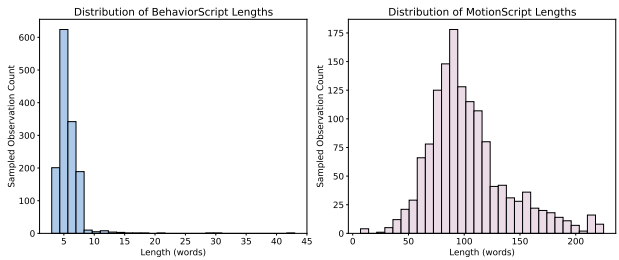

Figure 9. The Length statistics of sampled 1000 BehaviorScripts and corresponding MotionScripts.

The dataset features detailed statistics that emphasize its comprehensive scope. The sampled video clips predominantly range from 5 to 10 seconds in duration, ensuring that human actions are captured with sufficient temporal resolution. Annotated textual descriptions, including BehaviorScripts, average 10 to 15 words in length, offering concise yet informative summaries of the actions and context within the videos. MotionScripts, derived from these high-level descriptions, align closely with SMPL pose sequences, with segment lengths typically spanning 100–200 frames. Additionally, the dataset showcases linguistic diversity, as highlighted in the BehaviorScript word cloud, reflecting a wide array of actions, environments, and behavioral contexts. These characteristics make GBC-100k a versatile and robust dataset for advancing research in behavior modeling and multimodal learning.